# A Simplified Approach to Deep Learning for Image Segmentation

Ishtar Nyawĩra
Pittsburgh Supercomputing Center
Carnegie Mellon University
University of Pittsburgh
300 South Craig St.
Pittsburgh, PA  15213
1 (412) 268-4960
ishtarnyawira@pitt.edu

Kristi Bushman
Pittsburgh Supercomputing Center
Carnegie Mellon University
University of Pittsburgh
300 South Craig St.
Pittsburgh, PA  15213
1 (412) 268-4960
k.bushman@pitt.edu

## ABSTRACT

Leaping into the rapidly developing world of deep learning is an exciting and sometimes confusing adventure. All of the advice and tutorials available can be hard to organize and work through, especially when training specific models on specific datasets, different from those originally used to train the network. In this short guide, we aim to walk the reader through the techniques that we have used to successfully train two deep neural networks for pixel-wise classification, including some data management and augmentation approaches for working with image data that may be insufficiently annotated or relatively homogenous.

## 1 INTRODUCTION

In recent years, Convolutional Neural Networks (CNNs) have had extensive success in various visual recognition competitions. As a result, many new fields often involving data beyond that used for the earlier work on CNNs are attempting to make use of CNNs to automate repetitive tasks involving image data. With so many different frameworks, architectures, and data management techniques available, it can be difficult to know where to start when attempting to train a network with nontraditional data. In this paper, we will describe the approaches that we have used to successfully train two CNNs to segment neurons in high-resolution images of zebrafish larva brain tissue. We will also share some of the information and tips that we have learned along the way.

## 2 DATA MANIPULATION

Often, it is hard to gain access to enough data to train a deep network. One way to remedy this is by using data manipulation methods. Flipping and rotating images can grow a dataset by a factor of 8, assuming equal height and width. Translating images (i.e. shifting them within their borders) can grow a dataset by however many ways the data are translated. However, deep nets acquire more value from examples that vary more. Thus, flipping and rotating will likely help more than translating.

## 3 ATYPICAL DATA

Deep nets use edges and shapes to learn image data recognition and annotation. Data that is not annotated by fully outlining or filling in objects to be detected is not adequate for networks to learn from. Often, manually annotated data produces the best results, but at a significant cost of time. However, if adequate time is not available, one can place single-pixel click-points at the centers of each shape to be annotated then use flood-fill or region-growing (RG) operations. Both methods fill shapes from their centers to their borders. These methods are generally less successful than manually annotated data, but can be useful at the start as long as the objects being segmented have clearly defined borders and are relatively homogenous in color/intensity.

## 3.1 Flood-Fill & Region Growing Operations

The flood-fill algorithm [5] works to fill the shapes identified by centroids, from center to boundary. Assuming the images being used are grayscale, the algorithm is as follows:

1. *Make the image binary.*
2. *Extract an image skeleton, by reducing all the image's objects to lines (without altering its essential structure).*
3. *Apply a closing operation to ensure flood-fill leaking does not occur.*
4. *Flood-fill out from the click-point centers.*

The first step is necessary to reduce the number of pixel values (i.e., class values) to two: one for the neuron borders and the other for the background. The second step simplifies the complexity of the image so that the third step has cleaner lines to close and form defined shapes. The closing operation ensures that there are no gaps in the neuron borders for the flood-fill to leak out from. The final step does the actual filling of the neurons, from their click-points to their borders. See Appendix A for a visual guide.

This works to varying degrees of success and depends heavily on how well-defined those boundary lines are. If they are not dark or dense enough, the algorithm can overfill and spill beyond the edges of the shape (Fig. 1a). This can be prevented to some degree by increasing the radius of the structuring element for the closing operation. However, if increased too much, then small shapes are likely to be "ignored" and thus not filled at all or filled inadequately (Fig. 1b). Additionally, the larger the structuring element is, the more distorted the shapes may become.

The region growing algorithm [4] is a bit more robust and generally more thorough because it evaluates each pixel individually and then iteratively fills shapes based on those evaluations. Its algorithm is as follows:

1. *Get single-pixel seed (initial region).*
2. *Calculate the current region's mean intensity value.*
3. *Compare the intensity of all neighboring pixels to that of the current region's mean intensity.*
   a. *the similarity measure is simply the difference between the pixel & region's intensity.*
4. *Add the pixel with the best similarity (smallest difference) to the region.*
5. *Recalculate the region's mean intensity.*
6. *Repeat steps 3-5.*
7. *Stop when the similarity measure for all neighbors becomes larger than a given threshold.*

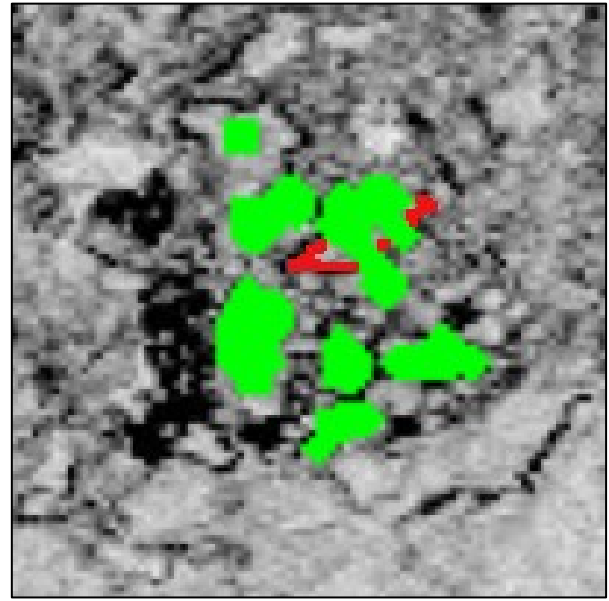

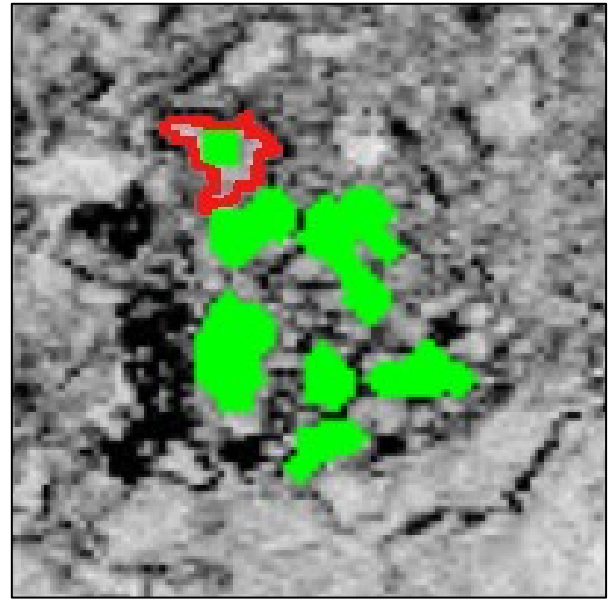

**Figures 1a & 1b: Flood-fill can spill over the boundaries of the neuron (top) or not fill the neuron to its boundaries (bottom).**

This method worked much better for our data, since the borders of the neurons in our images were not always clearly defined (Figs. 2a and 2b). This method is also much less parameter sensitive than the flood-fill method. However, if the shapes to be segmented are not homogenous in color/intensity, flood-fill might be a better option.

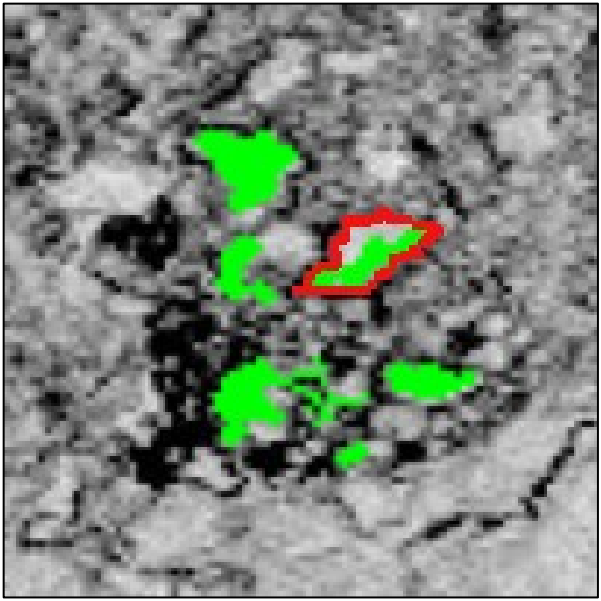

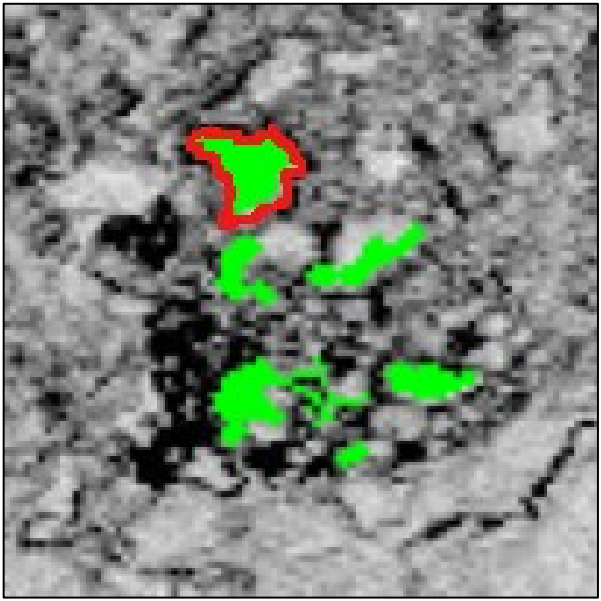

**Figures 2a & 2b: Region growing is much less prone to spilling over the boundaries of the neuron despite some trouble with very thin or pale membranes (top) and is better at filling the neuron to its boundaries (bottom).**

## 3.2  Deep Learning Frameworks & Available Networks

Several popular framework choices exist: Facebook's Caffe2, Google's TensorFlow, Theano, and Keras. Choosing a deep learning framework involves considering the merits and pitfalls of each. Caffe2's predecessor, Caffe, was developed by Berkeley Vision and Learning Center early in deep learning's renewed popularity. Thus, there is a substantial number of models built with Caffe and a large community using it. TensorFlow is more modular than Caffe, making it easier to significantly manipulate nets (an improvement adopted in Caffe2). It also has additional features, for example, TensorBoard, which helps summarize the kind of patterns picked up by networks and visualizes the learning process.

When choosing a pre-built network, it is important to consider the kind of data since each comes with its quirks. It is helpful to examine the kind of data being used by those who created the net. For example, medical image data, where classification is often binary and datasets lack variety, and road scene image data, which often involves multi-class classification and largely varied datasets, differ widely and require correspondingly different approaches.

## 4  EVALUATING RESULTS

To quantitatively evaluate our results, it is necessary to consider the nature of the data. If the data is medical image data, then generally it is going to be binary with one very large "background" class and one smaller class of interest (for example, a single cell type or small brain region). However, with pedestrian or street image data, there are often multiple classes of varying sizes (e.g., road, car, pedestrian, and sign). And still, one class can take up the majority of the image space.

To evaluate resulting predicted annotations, it is necessary to understand what constitutes a correctly annotated pixel and what does not. Common terminology with binary data (only two categories) includes the terms true positive (TP), true negative (TN), false positive (FP), and false negative (FN). True positives are pixels belonging to the category of interest (e.g. neuron) that are correctly annotated. True negatives are pixels of the other category (e.g., background) that are correctly annotated. False positives are pixels of the uninteresting category that are incorrectly annotated as pixels of interest (e.g., background pixels annotated as pixels belonging to a neuron). Finally, false negatives are pixels of interest incorrectly annotated as pixels of the other category (e.g. neuron pixels annotated as background pixels). When organized into a table, this is typically called a confusion matrix:

**Table 1: Simple Confusion Matrix**

|  | Predicted Positive | Predicted Negative |
|---|---|---|
| **Actual Positive** | TP | FN |
| **Actual Negative** | FP | TN |

When working with multiclass data that can share the image space, it's often popular to use both general accuracy (Equation 1) and the Intersection Over Union (IoU), also known as the Jaccard Index (Equation 2).

$$\text{ACC} = \frac{\text{TP+TN}}{\text{All}} \quad (1)$$

$$\text{JAC} = \frac{\text{TP}}{\text{TP+FP+FN}} \quad (2)$$

However, when working with binary data like medical image data, where class imbalance is common, it is best to use a metric that can prioritize the class of interest. Both the Area Under the Receiver Operating Characteristic Curve or ROC Curve (Equation 3) and the Cohen Kappa Coefficient (Equation 4) are good choices because they more accurately reflect the quality of segmentation for the smaller, more important class.

$$\text{AUROC} = 1 - \frac{\text{FPR}+\text{FNR}}{2} = 1 - \frac{1}{2}\left(\frac{\text{FP}}{\text{FP}+\text{TN}} + \frac{\text{FN}}{\text{FN}+\text{TP}}\right) \quad (3)$$

The AUROC is used to illustrate the probability that a randomly chosen positive pixel is correctly predicted to have a probability of being positive that is higher than that of a randomly chosen negative pixel. It uses two metrics that are actually sensitive to class imbalance and can be prone to overly penalizing errors in the smaller class [6]. Those two metrics are the TP rate, also called sensitivity or recall, and the FP rate, also called fall-out. As a result, networks trained with data that have high class imbalances, will return low sensitivity and fall-out values.

However, because the ROC plots these two metrics against one another, the focus remains on the relationship between these two values. Therefore, when the sensitivity increases, fallout decreases and vice versa. And the closer the AUROC is to 1, the more accurate the prediction model (Fig. 3).

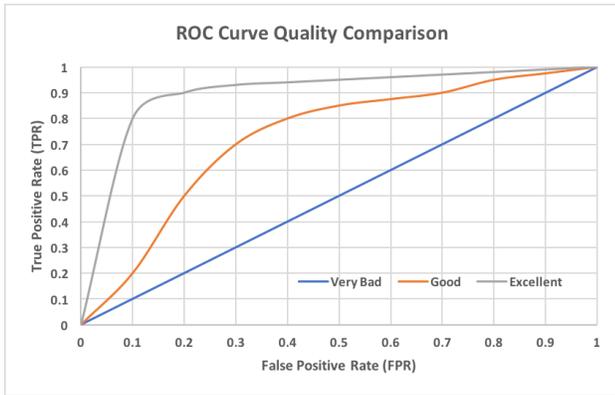

**Figure 3: Comparison of ROC Curves & their quality.**

The Cohen Kappa Coefficient on the other hand, measures the agreement between two samples, i.e. two pixels, while also accounting for chance agreement. Chance agreement is simply when two samples agree based on chance rather than based on the model's predictive powers. The Kappa measure takes the probability of agreement between classes (denoted below by Pa) and applies chance adjustment, by subtracting the hypothetical probability that a chance agreement happens (denoted by Pc), making this metric more robust [6].

$$\text{KAP} = \frac{\text{Pa}-\text{Pc}}{1-\text{Pc}} = \frac{\text{fa}-\text{fc}}{N-\text{fc}} \quad (4)$$

$$\text{fa} = \text{TP} + \text{TN} \quad (5)$$

$$\text{fc} = \frac{(\text{TN}+\text{FN})*(\text{TN}+\text{FP}) + (\text{FP}+\text{TP})*(\text{FN}+\text{TP})}{N} \quad (6)$$

Above, N is the total number of observations. However, because we have chosen to compute all of our confusion matrix elements as percentages, N is always equal to 1.

We should note that the Jaccard Index does eliminate TN from the numerator and denominator so that the focus remains on the pixels of the more important class. However, its inability to account for chance agreement makes it still sensitive to class imbalance [6].

## 4.1 Evaluating Metrics

With regards to evaluating the results of the Kappa measure and the AUROC, there isn't a lot of agreement. The choice is simpler for the AUROC, as it is generally evaluated using a traditional academic point system

**Table 2: Traditional Academic AUROC Curve Guidelines**

| Value | Level of Agreement |
|---|---|
| 0.50–0.59 | no agreement (F) |
| 0.60–0.69 | poor agreement (D) |
| 0.70–0.79 | fair agreement (C) |
| 0.80–0.89 | good agreement (B) |
| 0.90–1 | excellent agreement (A) |

Choosing a set of guidelines for the Kappa Coefficient is more challenging because both of the most popular guidelines are rather arbitrary. They are the Landis and Koch Guidelines [7] and the Fleiss Guidelines [8].

**Table 3: Landis and Koch Kappa Coefficient Guidelines**

| Value | Level of Agreement |
|---|---|
| < 0 | no agreement |
| 0–0.20 | slight agreement |
| 0.21–0.40 | fair agreement |
| 0.41–0.60 | moderate agreement |
| 0.61–0.80 | substantial agreement |
| 0.81–1 | almost perfect agreement |

**Table 4: Fleiss Kappa Coefficient Guidelines**

| Value | Level of Agreement |
|---|---|
| < 0.40 | poor agreement |
| 0.40–0.75 | fair to good agreement |
| > 0.75 | excellent agreement |

We focused on the Landis and Koch coefficients, as they provide a more detailed breakdown of agreement levels.

## 5 EXPERIMENT & DISCUSSION

In Tables 5 and 6 and Figs. 5 and 6 are some of the results we were able to acquire using two pre-built networks, SegNet [2] and UNet [3], and the flood-fill and region growing methods as well as some manual segmentation. Fig. 4 shows the corresponding original image and its true label.

**Table 5: UNet's Kappa Coefficient & AUROC Results w. Varying Net & Data Annotations**

|  | Kappa Coefficient | AUROC |
|---|---|---|
| **Flood-Fill (35,000 images)** | 0.674656 | 0.835366 |
| **Region Growing (35,000 images)** | 0.664252 | 0.775311 |
| **Manual Annotation (3,838 images)** | 0.508445 | 0.818312 |

**Table 6: SegNet's Kappa Coefficient & AUROC Results w. Varying Net & Data Annotations**

|  | Kappa Coefficient | AUROC |
|---|---|---|
| **Flood-Fill (35,000 images)** | 0.428534 | 0.816803 |
| **Region Growing (35,000 images)** | 0.557270 | 0.896916 |
| **Manual Annotation (3,838 images)** | 0.615110 | 0.965385 |

These results were gathered using models trained on 35,000 flood-filled training images, 35,000 region growing training images, and 3,838 manually annotated training images. The test set consisted of 50 randomly-selected images. The test set consisted of manually annotated images, since that is the closest that we have to a gold-standard annotation. The AUROC curve and Kappa measure were calculated using Equations 3 & 4 respectively.

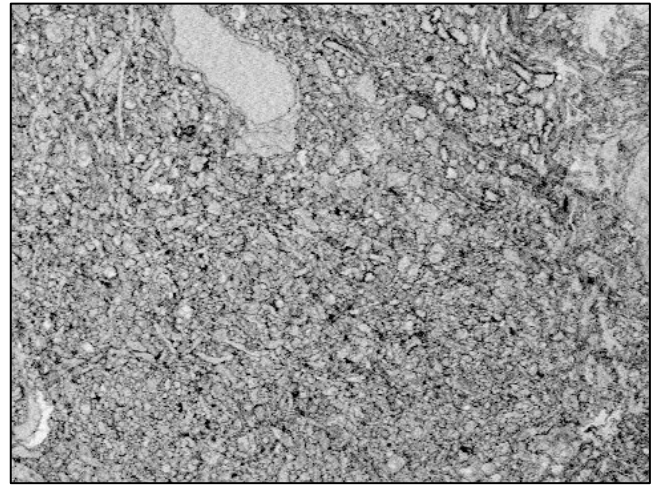
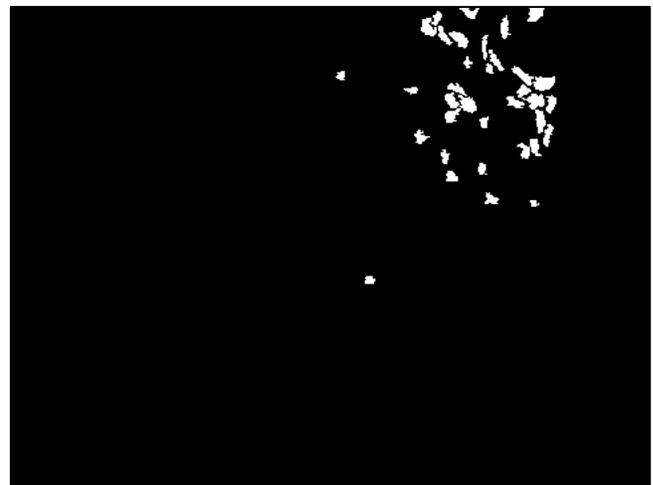

**Figures 4a & 4b: The original zebrafish slice (top) and true annotated label (bottom) from image 1930 of 4500. White spots are neuron segmentations.**

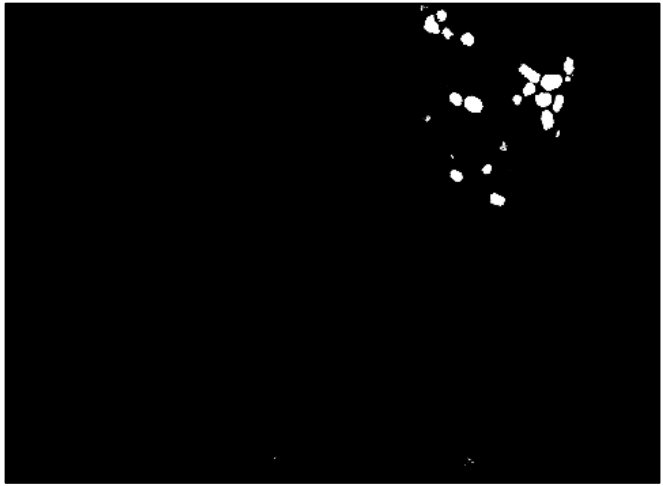
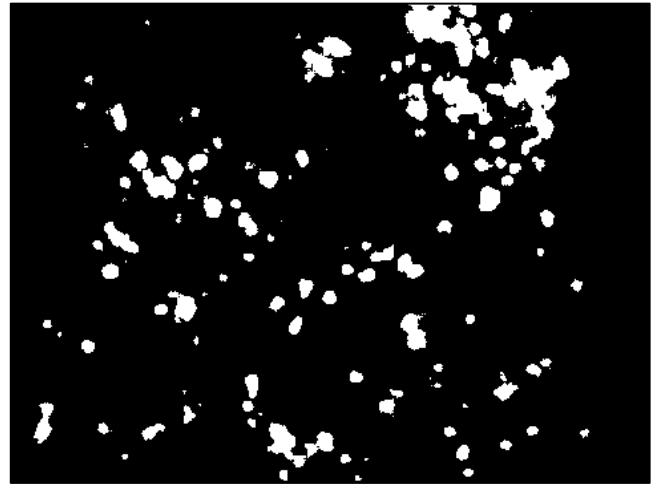
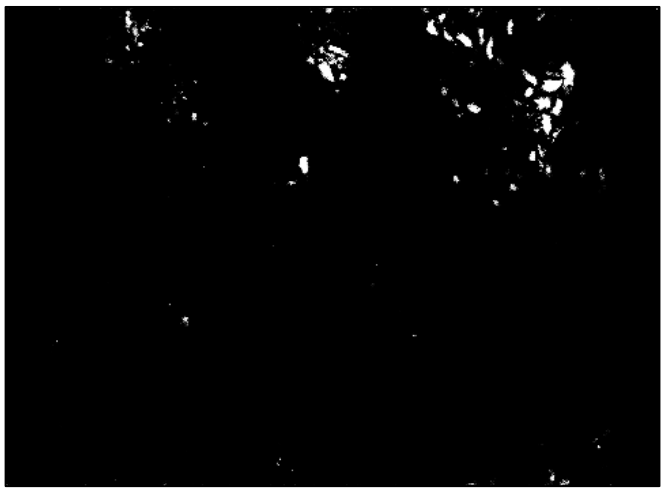
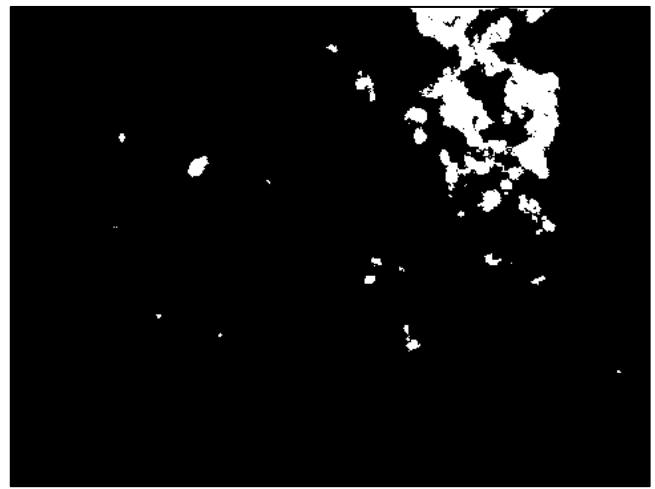
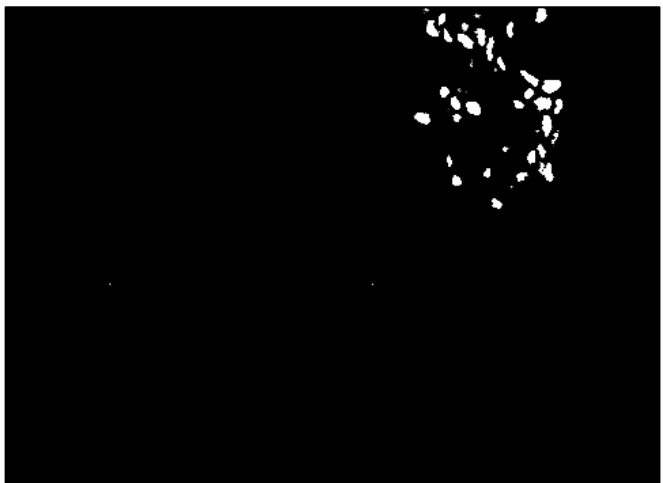
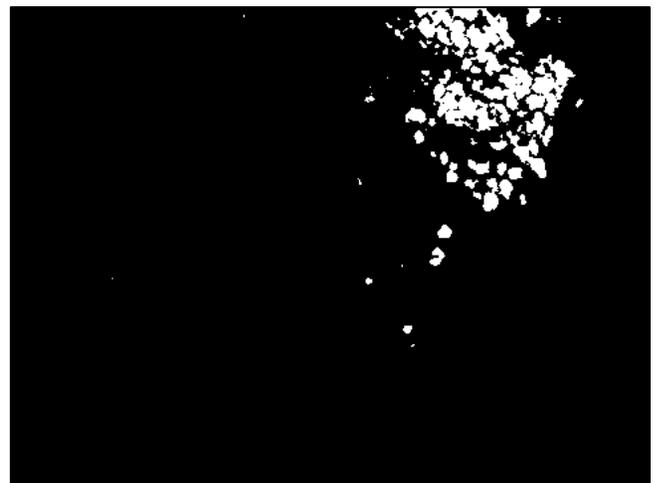

**Figures 5a & 5b & 5c:** UNet predicted results from image 1930 of 4500 using flood-fill (top), manual (middle) and region growing (bottom). White spots are neuron segmentations.

**Figures 6a & 6b & 6c:** SegNet predicted results from image 1930 of 4500 using flood-fill (top), manual (middle) and region growing (bottom). White spots are neuron segmentations.

The best quantitative results from SegNet come from our manually annotated images, even though these nets were trained with significantly less data. From these results, it appears as though it is more important for SegNet to have a small number of very well-annotated training data than to have an abundance of moderately well annotated data. Its AUROC results are significantly better than UNet's and but its Kappa Coefficient seems to be falling short of UNet's.

This suggests what we see in the qualitative results: that the more precise predicted segmentations come from UNet, as its better able to carefully define the contours of the neurons. This appears to be supported by UNet's generally higher Kappa measure results, which indicate that UNet is slightly better than SegNet at carefully segmenting individual neurons.

However, UNet's AUROC results are still lower than SegNet's and appear to be less closely linked to its Kappa measures. After some close evaluation, we were able to find that this was because UNet had a higher FN rate than SegNet. It was falsely identifying neuron pixels as background pixels more often. This was especially the case for the models trained on region growing and manually annotated data (Table 5). When factored into the AUROC (Equation 3), it would result in a significantly lower AUROC result.

**Table 7: FN Rate & FP Rate – UNet vs. SegNet**

|  | FNR | FPR |
|---|---|---|
| **UNet Flood** | 0.319044 | **0.010224** |
| **UNet RG** | 0.446302 | **0.003077** |
| **UNet Manual** | 0.334855 | 0.028521 |
| **SegNet Flood** | 0.322915 | **0.043478** |
| **SegNet RG** | 0.171410 | **0.034758** |
| **SegNet Manual** | 0.032981 | 0.036248 |

We presume that with more manually annotated training examples, SegNet's results will improve further and become more precise. On the other hand, UNet will likely require more investigation into class imbalance solutions to reduce the high FN Rate without subsequently increasing the FP Rate. Presently, we're working to manually annotate more images in order to have enough for even better results for both SegNet and UNet, as well as searching for better remedies for class imbalance.

## A  FLOOD-FILL ALGORITHM VISUAL GUIDE

This appendix functions as a visual guide to each step in the flood-fill process described in Section 3.1.

## A.1 Step-By-Step Visual Guide

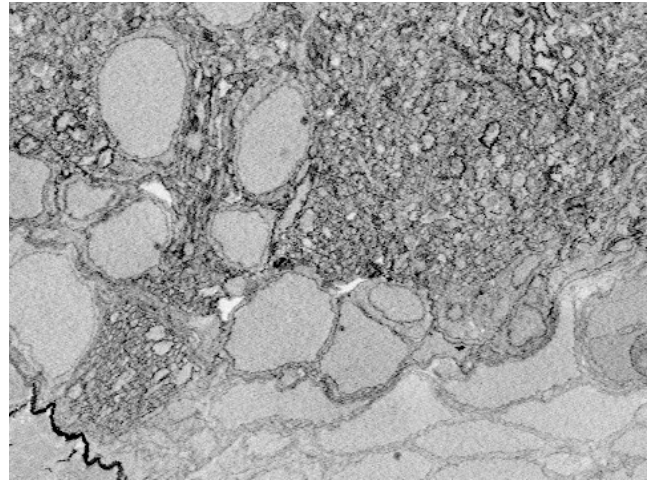

Original Image, Slice 1499 of 4900.

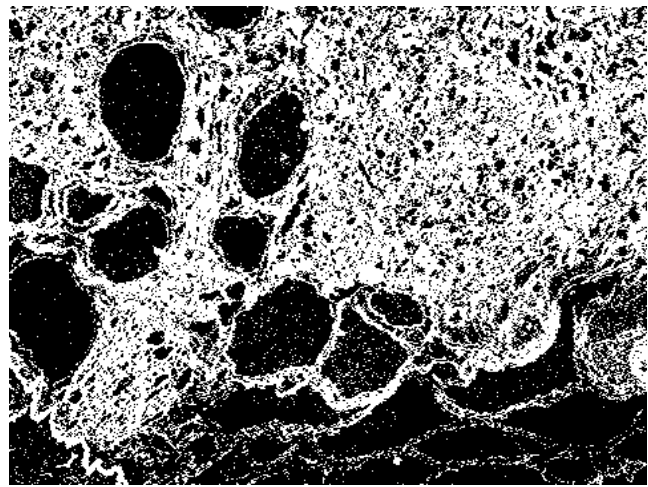

Step 1: Make the image binary.

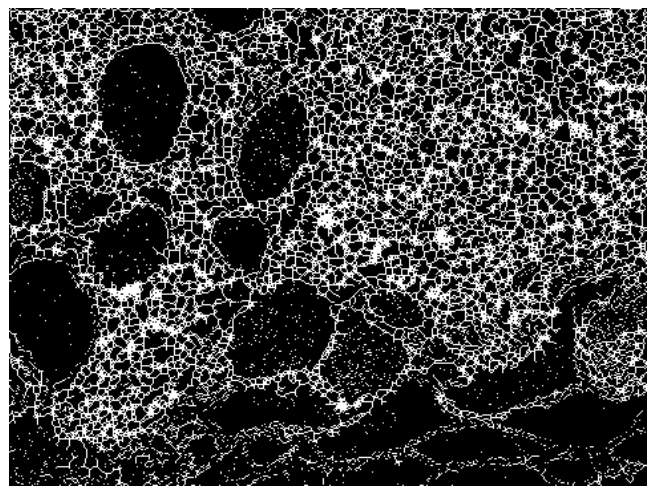

Step 2: Get the image skeleton.

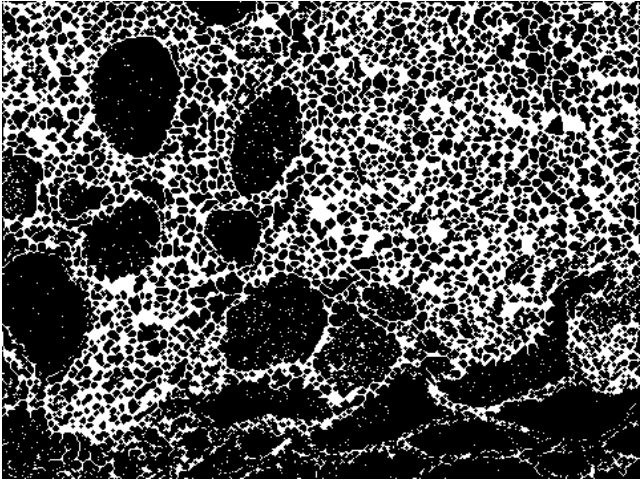

Step 3: Apply closing operation.

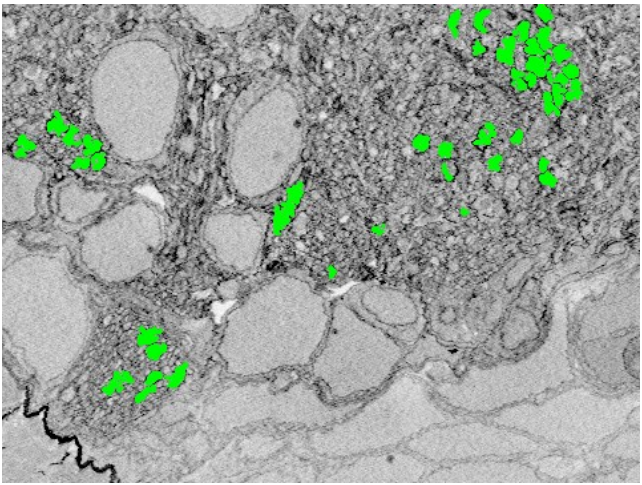

Step 4: Flood-fill the neurons from their centers to the borders defined in step 3.

## ACKNOWLEDGMENTS

This work used the Extreme Science and Engineering Discovery Environment (XSEDE), which is supported by National Science Foundation award number 1548562. Specifically, it used the Bridges system, which is supported by NSF award number 1445606, at the Pittsburgh Supercomputing Center (PSC). Support for student interns was provided by the Bridges project (NSF award 1445606) and the Commonwealth of Pennsylvania.

Special thanks to Nick Nystrom, Joel Welling, Paola Buitrago, John Urbanic, Arthur Wetzel, and Tamara Cherwin at PSC and Florian Engert, David Hildebrand, and their students at the Center for Brain Science at Harvard, who compiled and shared their zebrafish data.